\documentclass[letterpaper,10pt,conference]{ieeeconf}
\IEEEoverridecommandlockouts
\usepackage{graphicx}
\usepackage{tikz}      
\usepackage{amsmath,amssymb}
\usepackage{booktabs}
\usepackage{multirow}
\usepackage{cite}
\usepackage{algorithm}
\usepackage{algpseudocode}
\usepackage{siunitx}
\usepackage{array}
\usepackage{hyperref}
\hypersetup{hidelinks}
\usepackage{tikz}
\usetikzlibrary{positioning,calc,arrows.meta}
\graphicspath{{figures/}} 
\usepackage{indentfirst}  
\definecolor{softyellow}{RGB}{255,210,80}
\tikzset{
  flowarrow/.style={-{Latex[length=3.2mm,width=2.2mm]},line width=1.8pt,draw=softyellow,opacity=.95},
}
\newcommand{\leadit}[1]{\textit{#1:}\enspace}

\title{\LARGE \bf
Learning Adaptive Safety Margins for Visual Navigation\\
}
\author{Junyi Hu$^{1,\dagger}$, Shuaihang Yuan$^{1,\dagger,\ddagger}$, Geeta Chandra Raju Bethala$^{1}$, Anthony Tzes$^{1}$ and Yi Fang$^{1,*}$
\thanks{$^{1}$New York University Abu Dhabi, UAE. $^{\dagger}$Equal contribution. $^{\ddagger}$Project lead. $^{*}$Corresponding author: {\tt\small yfang@nyu.edu}}%
\thanks{Project page: \href{https://junyi2005.github.io/safety-critic/}{\tt\small junyi2005.github.io/safety-critic}}%
}
\begin{document}
\maketitle
\thispagestyle{empty}
\pagestyle{empty}

\begin{abstract}
Robots in cluttered indoor spaces often fail not because they cannot generate collision-free paths, but because a fixed safety margin is mis-calibrated: conservative margins cause detours and timeouts, while permissive margins lead to near-boundary shortcuts under perception bias.
Diffusion-based planners propose diverse trajectory candidates from egocentric RGB-D, yet reliable selection remains the bottleneck.
We propose a context-conditioned safety critic that learns an adaptive clearance preference for ranking diffusion proposals, decomposed into three complementary terms: (i) a safety term with a clearance-budget penalty and a control-barrier-function residual for waypoint- and transition-wise safety, (ii) an efficiency term combining a smoothness penalty with a safety-gated detour-ratio penalty that avoids detours without incentivizing risky shortcuts, and (iii) a distance-constraint matching term that anchors the learned budget to realized ESDF clearances to prevent margin collapse.
We train the critic with privileged ESDF geometry in simulation and distill it into a perception-only selector via a two-stage teacher--student procedure.
On PointGoal navigation in HM3D and MP3D, including cross-dataset transfer, our method achieves the highest success rate (SR) and success weighted by path length (SPL) among strong diffusion, optimization, and RL baselines.
Trained purely in simulation, it transfers to a Unitree G1 humanoid and navigates cluttered indoor scenes without task-specific tuning.
\end{abstract}

\begin{figure}[t]
\centering
\includegraphics[width=0.72\linewidth]{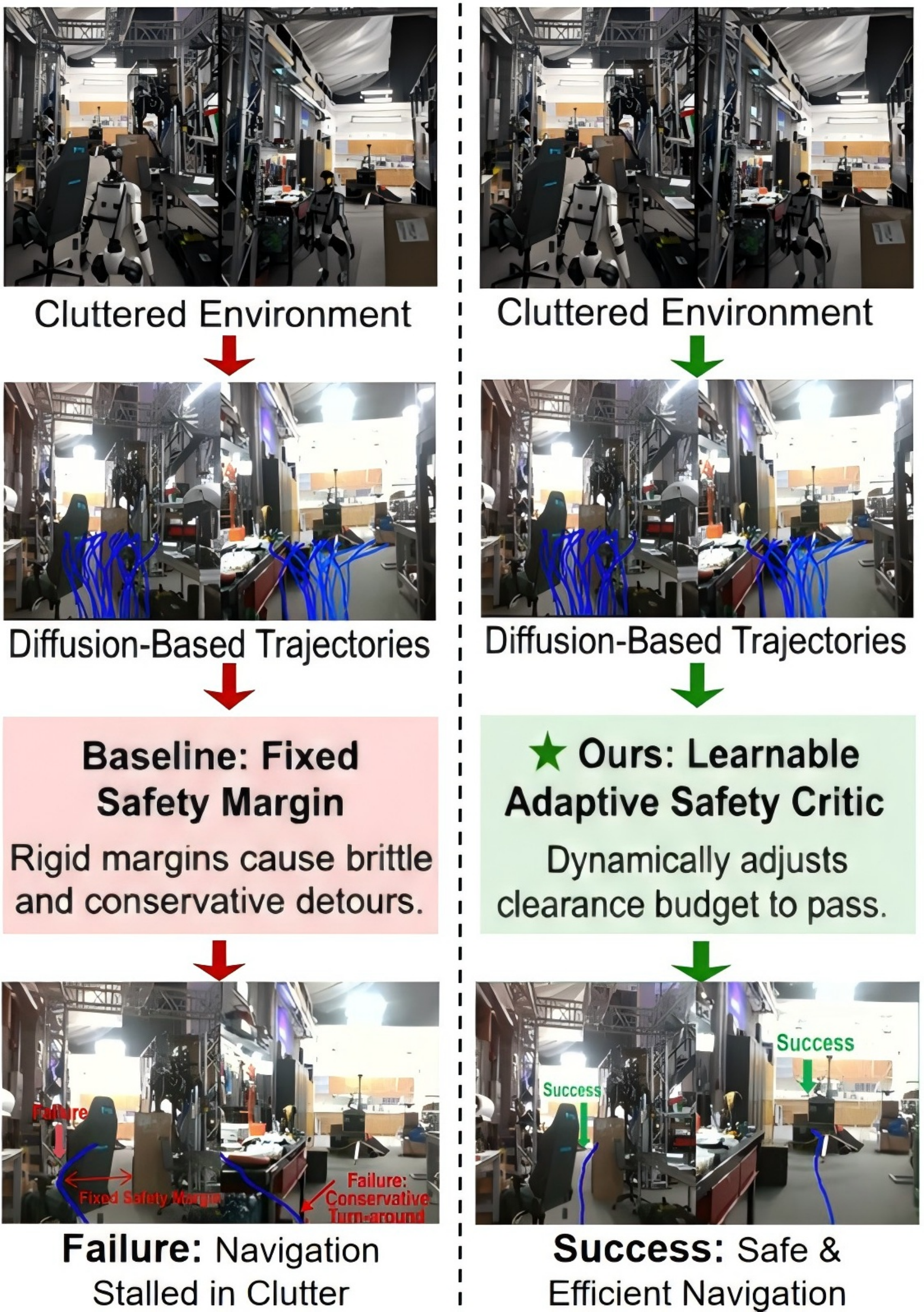}
\caption{Comparison between a fixed-margin critic and our adaptive safety critic. Fixed margins often lead to conservative detours or brittle near-boundary choices, while our critic learns a trajectory-dependent clearance budget for robust navigation in clutter.}
\label{fig:ind}
\end{figure}
\section{Introduction}
Mobile robots are increasingly deployed in human-centric indoor environments such as homes, hospitals, and labs, where success depends not only on geometric feasibility but on maintaining sufficient clearance while executing a trajectory.
Crucially, the required clearance is context dependent: local clutter, maneuvering demands (e.g., sharp turns in narrow passages), and momentary perception reliability can make the same measured distance safe in one situation and risky in another. This is amplified in vision-based navigation, where depth and egomotion estimates vary with viewpoint, motion, and sensing artifacts, so minimum measured clearance is a noisy proxy for executability.
As a result, hard selection rules with a globally fixed clearance threshold oscillate between two failure modes: overly conservative behavior in open areas (detours and timeouts) and brittle behavior in dense clutter (risky shortcuts under perception bias).

Diffusion-based policies model multi-modal trajectory distributions conditioned on observations, enabling diverse, feasible proposals and a practical generate--select paradigm \cite{chi2023diffusionpolicy}.
However, the selector is often the bottleneck: many diffusion navigation pipelines still rank candidates using analytic critics with hand-tuned, globally fixed safety margins \cite{navdp,tdiff2024}, which cannot be simultaneously conservative enough for tight clutter and permissive enough to avoid detours in open space.
This motivates a selector whose notion of ``safe distance'' is learned and adapts its safety--efficiency trade-off to the current scene.

In this work, we propose a context-conditioned safety critic for diffusion-based visual navigation that replaces fixed-margin ranking with a learnable clearance preference, structured into three complementary terms:
(i) \emph{Safety}: penalizes clearance-budget violations and incorporates a control-barrier-function residual for transition-wise regularization;
(ii) \emph{Efficiency}: combines a smoothness penalty with a safety-gated detour-ratio penalty that discourages detours primarily when clearance is available; and
(iii) \emph{Balance}: a distance-constraint matching term that aligns the learned budget with realized clearances, preventing degenerate calibration.
We train a teacher critic using ESDF-derived clearance supervision and hard non-expert trajectories under matched start--goal conditions, then distill it into a student selector that predicts critic scores from egocentric RGB-D observations, enabling deployment without privileged geometry.

We evaluate PointGoal navigation on HM3D and MP3D, in-domain and under cross-dataset transfer, where our method achieves the highest SR and SPL among strong diffusion, optimization-based, and reinforcement-learning baselines. We further demonstrate sim-to-real transfer on a Unitree G1 humanoid in cluttered indoor scenes without task-specific tuning, and our ablations isolate the contribution of each critic component.

\section{Related Work}
\label{sec:Related Work}
End-to-end visual navigation closes the perception--action loop by mapping egocentric observations to low-level actions or short-horizon motion segments, reducing latency and error propagation across perception, mapping, and planning.
Large-scale simulators such as Habitat provide standardized embodied tasks, where deep RL agents (e.g., DD-PPO) set strong PointGoal baselines \cite{wijmans2019dd}, and hybrid systems such as Active Neural SLAM add explicit spatial memory via learned mapping and hierarchical policies over a top-down map \cite{ans2020}.

Recent generative decision-making models multi-modal distributions over action or trajectory sequences, providing a principled way to sample diverse candidates under conditioning \cite{chi2023diffusionpolicy,he2023diffusionplanner}.
Beyond learned generators, iPlanner couples perception with a differentiable cost map and bi-level optimization to produce collision-free paths from a single depth measurement \cite{iplanner2024}, and VLFM scores candidate frontiers with a pre-trained vision-language model for zero-shot navigation \cite{vlfm2024}.
On the model-based side, geometry-driven local planning remains competitive when paired with high-quality distance fields such as Voxblox \cite{voxblox2017}, while EGO-Planner performs real-time gradient-based replanning without explicit ESDF construction \cite{zhou2020ego}.
Despite their differences, these systems ultimately depend on a hand-crafted scoring rule---often a fixed clearance margin---to arbitrate among candidates, which can yield unsafe shortcuts under perception bias or overly conservative detours in clutter.

\section{Methodology}
\label{sec:method}

\leadit{Task Formulation}Given RGB-D observations, the robot's current pose $\mathbf{p}^{R}_t=(x_0,z_0,\theta_0)\in\mathbb{R}^3$, a goal pose $\mathbf{p}^{G}_t$, and an obstacle set $\mathcal{Q}_{\text{o}}$, the objective is to compute a finite-horizon trajectory $\tau_t=\{\mathbf{p}_j\}_{j=0}^{T}$ that starts at $\mathbf{p}_0=\mathbf{p}^{R}_t$, reaches $\mathbf{p}^{G}_t$, and is collision-free.
Let $\pi(\mathbf{p})=(x,z)$ be the 2D planar projection.
We parameterize the trajectory by ego-motion offsets $\mathbf{u}_j=(\Delta x_j,\Delta z_j,\Delta \theta_j)$ for $j=1,\dots,T$ (lateral/longitudinal displacements and heading offset) in the initial egocentric frame anchored at $\mathbf{p}^{R}_t$, and reconstruct poses by $\mathbf{p}_j=\mathbf{p}_0\oplus \mathbf{u}_j$.
The task is to find $\{\mathbf{u}_j\}_{j=1}^{T}$ such that each segment from $\mathbf{p}_{j-1}$ to $\mathbf{p}_j$ avoids $\mathcal{Q}_{\text{o}}$ while driving the robot from $\mathbf{p}^{R}_t$ to $\mathbf{p}^{G}_t$.

\begin{figure*}[!t]
  \centering
  \includegraphics[width=\textwidth]{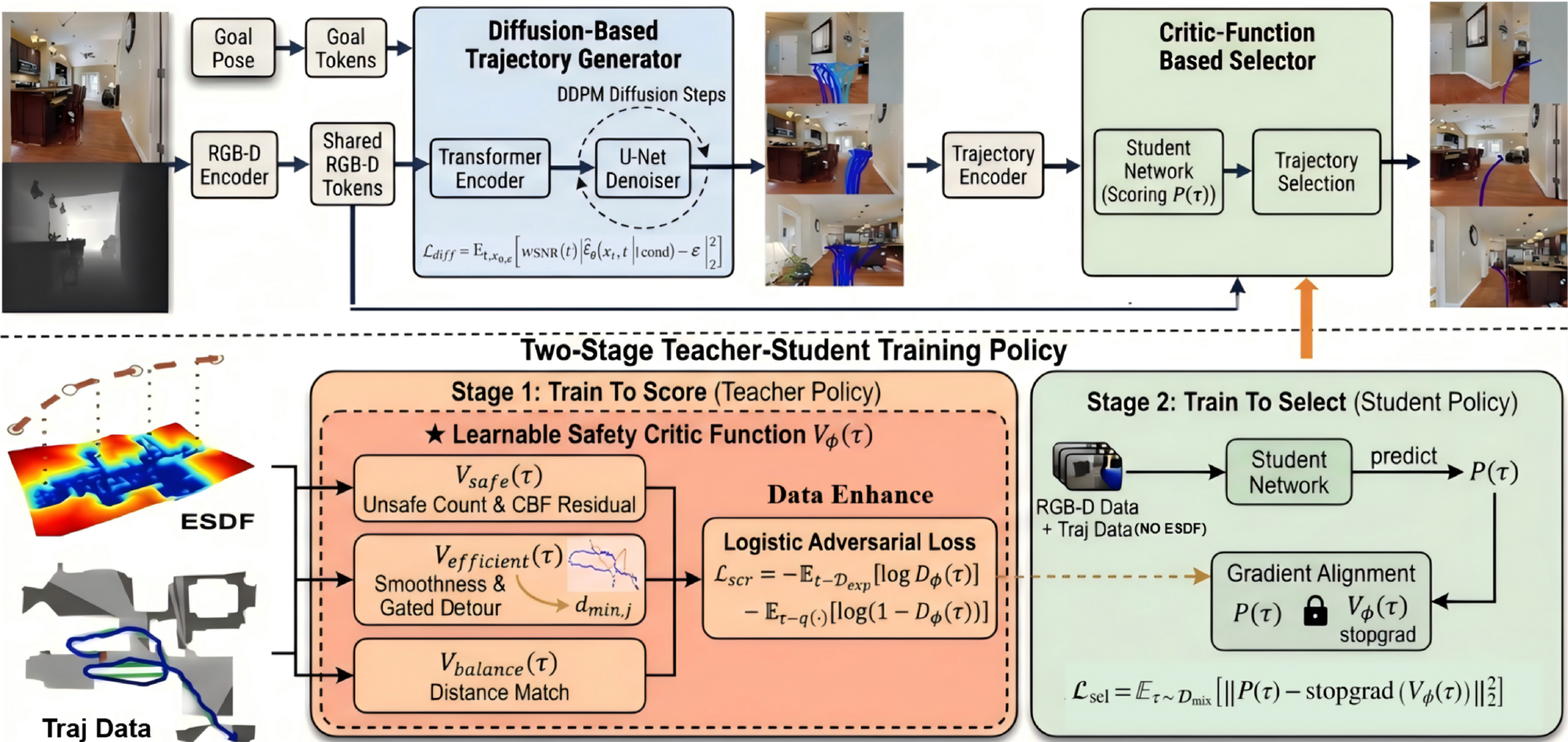}
  \caption{Architecture of our system. During inference, the diffusion policy samples candidate trajectories from RGB-D observations and the goal. ESDF is used only offline to supervise the teacher during training, while the deployed selector runs without map building. We first train a learnable safety critic with ESDF-based supervision, then distill it into a selector with RGB-D observations and candidate trajectories input via teacher--student training.}
  \label{fig:overview}
\end{figure*}

\leadit{System Overview}Our policy follows a generate--select pipeline (Fig.~\ref{fig:overview}): a diffusion generator produces diverse candidates from RGB-D observations and an optional goal token, and a learnable safety critic with context-dependent margins selects among them. The selector is trained with a two-stage teacher--student scheme (Sec.~\ref{sec:train}), where ESDF is used only offline to supervise the teacher.

\subsection{Diffusion Trajectory Generator}
\label{sec:diffusion}
We follow the diffusion trajectory generation setup used in NavDP \cite{navdp} and T-diff \cite{tdiff2024}.
A shared transformer encoder consumes the RGB-D observations and an optional goal token to produce a compact context, and a diffusion head generates $K$ trajectories ($K=16$ by default) without privileged geometric inputs; diversity is obtained by sampling different noise seeds under the same denoiser and scheduler.

During training, we optimize a conditional U-Net denoiser to predict injected Gaussian noise at randomly sampled diffusion steps under a DDPM scheduler \cite{ho2020ddpm} with a squared-cosine noise schedule, fusing normalized last-step action deltas with timestep, goal, and RGB-D embeddings under a causal target mask.
Let $\beta_t\!\in\!(0,1)$ be the per-step noise rate, $\alpha_t=1-\beta_t$, and $\bar{\alpha}_t=\prod_{s=1}^{t}\alpha_s$ with $\bar{\alpha}_0=1$.
We use stabilized SNR weighting computed from $\bar{\alpha}_t$,
\begin{equation}
\label{eq:snr}
\mathrm{SNR}(t)=\frac{\bar{\alpha}_t}{1-\bar{\alpha}_t},\qquad
w_{\mathrm{SNR}}(t)=\min\!\big(\mathrm{SNR}(t),\,\tau_{\mathrm{SNR}}\big),
\end{equation}
where $\tau_{\mathrm{SNR}}>0$ avoids over-weighting very early steps.
The diffusion loss is
\begin{equation}
\label{eq:ldiff}
\mathcal{L}_{\mathrm{diff}}
~=~\mathbb{E}_{t,\mathbf{x}_0,\epsilon}\!\left[
w_{\mathrm{SNR}}(t)\,\big\|\hat{\epsilon}_\theta(\mathbf{x}_t,t\,|\,\text{cond})-\epsilon\big\|_2^2
\right],
\end{equation}
optimized with AdamW, cosine annealing, automatic mixed precision, and gradient norm clipping; ESDF is not used in training or inference of the generator.

At inference, since the offsets $(\Delta x_j,\Delta z_j,\Delta \theta_j)$ are expressed in the initial egocentric frame, a single rotation $R(\theta_0)$ reconstructs world-frame waypoints,
\begin{equation}
\label{eq:reconstruct-start}
\theta_{j}=\theta_{0}+\Delta \theta_{j},\qquad
\begin{bmatrix}x_{j}\\ z_{j}\end{bmatrix}
=
\begin{bmatrix}x_{0}\\ z_{0}\end{bmatrix}
+
R(\theta_{0})\begin{bmatrix}\Delta x_{j}\\ \Delta z_{j}\end{bmatrix}.
\end{equation}
Control commands are obtained from finite differences $r_{j}=(x_{j+1}-x_{j},\, z_{j+1}-z_{j},\, \theta_{j+1}-\theta_{j})$, wrapping the heading increment into $(-\pi,\pi]$ and computing speed and angular velocity from $r_{j}$.

\subsection{Learnable Safety Critic}
\label{sec:fused_critic}
We use the collected trajectories in waypoint form $\{(x_j,\, z_j,\, \theta_j)\}_{j=0}^{T}$ to train the safety-critic value function, with the planar position $\pi(\mathbf{p}_j)=(x_j,z_j)$ queried in the ESDF as the key variable for scoring.
Our safety critic is decomposed into three components in~\eqref{eq:v_ours_sum}: a safety term $V_{\mathrm{safe}}$, an efficiency term $V_{\mathrm{efficient}}$, and a balancing term $V_{\mathrm{balance}}$, detailed below.

Rather than treating $d_{\min,j}$ as a globally shared parameter, we use a context-conditioned margin head $q_{\eta}$ to predict a time-varying safety budget,
\begin{equation}
\label{eq:dmin_context}
d_{\min,j}=d_{\mathrm{safe}}+\operatorname{softplus}\bigl(q_{\eta}(\mathbf{f}_j)\bigr),
\end{equation}
where $d_{\mathrm{safe}}=0.1$\,m is a fixed physical safety floor, and $\mathbf{f}_j$ is a lightweight context feature (local clearance $d_j$, a finite-difference ESDF gradient magnitude, and normalized step index $j/T$) fed to a 2-layer MLP $q_\eta$.
This guarantees $d_{\min,j}\ge d_{\mathrm{safe}}$ and makes the clearance budget trajectory- and geometry-dependent during teacher training.

Let $d_j=\mathrm{ESDF}(\pi(\mathbf{p}_j))$ be the distance between the $j$-th waypoint and the obstacle in ESDF, and let $\Delta^2 \pi(\mathbf{p}_j)=\pi(\mathbf{p}_{j+1})-2\pi(\mathbf{p}_j)+\pi(\mathbf{p}_{j-1})$ denote the discrete second difference used for the smoothness term.
We define the safe set $\mathcal{C}=\{p\,|\,h(p)\ge 0\}$ with barrier function $h(p)=d(p)-d_{\mathrm{safe}}$.
For a discrete trajectory, the Control Barrier Function (CBF) residual is $r^{\mathrm{cbf}}_{j}=(1-\rho)h(\mathbf{p}_j)-h(\mathbf{p}_{j+1})$ for $j=0,\dots,T-1$, where $\rho=0.1$ controls conservativeness, and the CBF critic loss is
\begin{equation}
\mathcal{L}_{\mathrm{cbf}}(\tau)=\sum_{j=0}^{T-1}\big[r^{\mathrm{cbf}}_{j}\big]_+.
\end{equation}
\begin{equation}
\label{eq:v_safe}
V_{\mathrm{safe}}(\tau)=
-\sum_{j=0}^{T}\mathbb{I}\!\bigl(d_j< d_{\min,j}\bigr)
-\lambda_{\mathrm{cbf}}\,\mathcal{L}_{\mathrm{cbf}}(\tau).
\end{equation}
The safety term $V_{\mathrm{safe}}(\tau)$ encodes two complementary mechanisms: the unsafe-count component penalizes violations of the learned budget $d_{\min,j}$ at each waypoint, while the CBF critic loss enforces transition-wise safety-set invariance with respect to the fixed floor $d_{\mathrm{safe}}$, so the critic prefers candidates that remain within a physically meaningful safety set while the budget adapts across contexts.

The gated detour-ratio loss encourages efficiency without requiring explicit start-to-goal or geodesic shortest-path supervision.
Let $L_{\mathrm{path}}(\tau)=\sum_{j=1}^{T}\|\pi(\mathbf{p}_j)-\pi(\mathbf{p}_{j-1})\|_2$ be the accumulated path length and $D_{\mathrm{chord}}(\tau)=\|\pi(\mathbf{p}_T)-\pi(\mathbf{p}_0)\|_2$ the start-to-endpoint chord length.
We define the detour-ratio loss as a hinge on the relative redundancy of the candidate path length,
\begin{equation}
\label{eq:detour_ratio_loss}
\mathcal{L}_{\mathrm{detour}}(\tau)=\left[\frac{L_{\mathrm{path}}(\tau)}{D_{\mathrm{chord}}(\tau)+\varepsilon}-1\right]_+,
\end{equation}
where $\varepsilon>0$ is a small constant for numerical stability.
To prevent the efficiency objective from incentivizing risky shortcuts in near-contact regions, we couple the detour-ratio penalty with a safety gate derived from ESDF clearance and the learnable budget, using the per-step gate weight $w_j=\sigma\!\left(\kappa\left(d_j-d_{\min,j}\right)\right)$, where $\sigma(\cdot)$ is the sigmoid and $\kappa>0$ controls sharpness. Aggregating these into a trajectory-level gate $\bar{w}(\tau)=\frac{1}{T+1}\sum_{j=0}^{T}w_j$ gives the gated detour-ratio loss $\mathcal{L}^{\mathrm{gate}}_{\mathrm{detour}}(\tau)=\bar{w}(\tau)\,\mathcal{L}_{\mathrm{detour}}(\tau)$.
\begin{equation}
\label{eq:v_efficient}
V_{\mathrm{efficient}}(\tau)=
-\beta\sum_{j=1}^{T-1}\left\|\Delta^2 \pi(\mathbf{p}_j)\right\|_2
-\mu\,\mathcal{L}^{\mathrm{gate}}_{\mathrm{detour}}(\tau).
\end{equation}
The efficiency term $V_{\mathrm{efficient}}(\tau)$ combines a curvature-based smoothness penalty with the gated detour-ratio penalty, where $\bar{w}(\tau)$ emphasizes the penalty when clearance stays above the budget and weakens it near the safety boundary.

\begin{equation}
\label{eq:v_balance}
V_{\mathrm{balance}}(\tau)=
-\psi\sum_{j=0}^{T}\left(d_j-d_{\min,j}\right)^{2}.
\end{equation}
The balancing term $V_{\mathrm{balance}}(\tau)$ couples the learned budget to the realized clearance and stabilizes the trade-off: $d_{\min,j}$ is pulled toward the lower bound in tight corridors so feasible candidates are not rejected and increases in open areas, while reducing $d_{\min,j}$ without increasing clearance lowers the score, preventing margin collapse.

Overall, we define our safety-critic as:
\begin{equation}
\label{eq:v_ours_sum}
V_{\mathrm{ours}}(\tau)=V_{\mathrm{safe}}(\tau)+V_{\mathrm{efficient}}(\tau)+V_{\mathrm{balance}}(\tau).
\end{equation}
We learn the unconstrained weight variables $\tilde{\mathbf{w}}=[\tilde{\beta},\tilde{\lambda}_{\mathrm{cbf}},\tilde{\mu},\tilde{\psi}]^\top$ and the margin head $q_\eta(\cdot)$, enforcing nonnegative penalty weights via the reparameterization $\mathbf{w}=\operatorname{softplus}(\tilde{\mathbf{w}})$ with $\mathbf{w}=[\beta,\lambda_{\mathrm{cbf}},\mu,\psi]^\top$.
Since the data are collected in simulation with safe kinematic settings, velocity and acceleration already lie within safe ranges, so we add no CBF constraints on them.
\begin{table*}[t]
\centering
\caption{Results for Three Simulation Tests.}
\label{tab:sim}
\setlength{\tabcolsep}{10.5pt}
\renewcommand{\arraystretch}{1.05}
\begin{tabular}{lccc ccc ccc}
\toprule
& \multicolumn{3}{c}{HM3D results} & \multicolumn{3}{c}{MP3D results} & \multicolumn{3}{c}{Cross-dataset transfer} \\
\cmidrule(lr){2-4}\cmidrule(lr){5-7}\cmidrule(lr){8-10}
Method & Successes & SR $\uparrow$ & SPL $\uparrow$
       & Successes & SR $\uparrow$ & SPL $\uparrow$
       & Successes & SR $\uparrow$ & SPL $\uparrow$ \\
\midrule
NavDP                 & 213/300 & 0.710 & 0.529 & 188/300 & 0.627 & 0.431 & 157/300 & 0.523 & 0.349 \\
iPlanner              & 196/300 & 0.653 & 0.573 & 181/300 & 0.603 & 0.492 & 171/300 & 0.570 & 0.436 \\
PointNav              & 164/300 & 0.547 & 0.406 & 117/300 & 0.390 & 0.251 & 102/300 & 0.340 & 0.216 \\
ViPlanner             & 205/300 & 0.683 & 0.508 & 173/300 & 0.577 & 0.407 & 166/300 & 0.553 & 0.413 \\
\textbf{Ours}         & \textbf{235/300} & \textbf{0.783} & \textbf{0.611} & \textbf{204/300} & \textbf{0.680} & \textbf{0.523} & \textbf{179/300} & \textbf{0.597} & \textbf{0.465} \\
\bottomrule
\end{tabular}
\end{table*}
\subsection{Selector Training}
\label{sec:train}
The selector is trained in two stages: the first trains the safety critic as a teacher (learnable penalty weights and margin head) using ESDF-based supervision; the second trains the selector as a student to imitate the teacher for scoring and selection from RGB-D observations and candidate trajectories, enabling deployment without privileged geometry.

We collect a dataset of smooth paths and paired RGB-D observations from 3D scenes by sampling random endpoints, planning via A*, and applying cubic spline interpolation.
Each trajectory contains over 80 waypoints; we extract five overlapping sliding windows per trajectory and resample each to $T=24$ waypoints, and compute a high-resolution ESDF of the navigable space from the scene voxel map.
During teacher training, we also construct non-expert trajectories directly in the global ESDF space under matched start--goal configurations, providing continually refreshed challenging negatives without handcrafted disturbance rules.

We denote the trajectory score in~\eqref{eq:v_ours_sum} by $V_\phi(\tau)$.
Since $V_\phi(\tau)$ is a sum of penalty-style terms and is non-positive by design, we introduce a nonnegative cost $C_\phi(\tau)=-V_\phi(\tau)\ge 0$.
To train the teacher in an adversarial classification form, we use an affine-calibrated discriminator head
\begin{equation}
\label{eq:scr_disc_def}
D_\phi(\tau)=\sigma\!\bigl(-a\,C_\phi(\tau)+b\bigr),
\end{equation}
where $\sigma(\cdot)$ is the sigmoid, $a=\operatorname{softplus}(\hat{a})>0$ is a scale, and $b$ is a calibration bias, equivalently $D_\phi(\tau)=\sigma\!\bigl(aV_\phi(\tau)+b\bigr)$.
Let $\mathcal{D}_{\mathrm{exp}}$ denote expert sub-trajectories obtained from A* planning and spline smoothing.
For each, we generate ESDF-conditioned non-expert candidates from a proposal distribution $q(\tau\,|\,\mathbf{p}^{R}_t,\mathbf{p}^{G}_t,\mathrm{ESDF})$ that depends only on global geometry: we repeatedly run A* with randomized edge costs between the same start and goal, reject any path violating $d<d_{\mathrm{safe}}$, and apply the same resampling to obtain length-$T$ sequences.
The adaptive margin sequence $\{d_{\min,j}\}_{j=0}^{T}$ is obtained by a single forward pass of $q_\eta$, without per-trajectory online optimization.
The teacher is trained to assign larger discriminator probabilities to expert trajectories and smaller ones to non-expert trajectories using the logistic adversarial loss:
\begin{equation}
\label{eq:scr_adv}
\begin{aligned}
\mathcal{L}_{\mathrm{scr}}
&=
-\mathbb{E}_{\tau\sim\mathcal{D}_{\mathrm{exp}}}\!\left[\log D_\phi(\tau)\right] \\
&\quad
-\mathbb{E}_{\tau\sim q(\cdot\,|\,\mathbf{p}^{R}_t,\mathbf{p}^{G}_t,\mathrm{ESDF})}\!\left[\log\!\bigl(1-D_\phi(\tau)\bigr)\right].
\end{aligned}
\end{equation}

Instead of a hard-margin ranking loss, this learns a decision boundary between expert- and non-expert-like trajectories under matched start--goal contexts in the same ESDF, with gradients backpropagated to $\boldsymbol\phi$ and calibration parameters $a,b$.

After the first stage, we train the student selector to imitate the teacher. Each candidate is encoded as a token from its ego-motion offsets $\mathbf{u}_j$; goal tokens are omitted, as the teacher's efficiency term relies solely on trajectory geometry. Together with RGB-D observations, the selector predicts a score trained to regress the teacher scores by minimizing
\begin{equation}
\label{eq:loss_selector}
\mathcal{L}_{\mathrm{sel}}
=
\mathbb{E}_{\tau\sim\mathcal{D}_{\mathrm{mix}}}\!\left[\bigl\|P(\tau)-\operatorname{stopgrad}\!\bigl(V_\phi(\tau)\bigr)\bigr\|_2^2\right],
\end{equation}
where $P(\cdot)$ is the selector prediction and $\mathcal{D}_{\mathrm{mix}}$ mixes expert sub-trajectories from $\mathcal{D}_{\mathrm{exp}}$ and diffusion-generated candidates sampled under the same RGB-D observations used at inference.
At inference, the selector scores diffusion candidates from RGB-D observations and trajectory tokens without ESDF reconstruction and selects the most suitable trajectory.
\section{Experiments}

\label{sec:eval}

\subsection{Simulation Experiments}
\label{sec:simtest}
We benchmark PointGoal navigation on HM3D~\cite{hm3d} and MP3D~\cite{mp3d}. All methods use the \emph{same} start--goal pair per episode with a 500-step timeout, and succeed once the agent reaches the goal within the standard PointGoal radius. We compare four baselines: NavDP~\cite{navdp} (diffusion generator with a fixed-constraint selector), iPlanner~\cite{iplanner2024} (differentiable trajectory optimizer), ViPlanner~\cite{roth2024viplanner} (learned visual module with a classical backbone), and PointNav~\cite{wijmans2019dd} (end-to-end RL), matching action frequency, observation format, kinematic constraints, and termination rules. PointNav is trained in Habitat-Sim; NavDP and iPlanner are retrained on HM3D-converted data using official code. We evaluate HM3D$\rightarrow$HM3D, MP3D$\rightarrow$MP3D, and HM3D$\rightarrow$MP3D with 300 episodes per policy and identical episode sets, reporting SR and SPL.
The results are in Table~\ref{tab:sim}.

On HM3D, our method achieves the best SR \textbf{0.783} (\textbf{+7.3} over NavDP, \textbf{+23.6} over PointNav) and the best SPL \textbf{0.611}, indicating the learnable critic both avoids risky choices and favors efficient routes. On MP3D it again ranks first with SR \textbf{0.680} (\textbf{+5.3} over NavDP) and SPL \textbf{0.523}; SPL drops for all methods due to MP3D's longer corridors and denser clutter, yet our learnable $d_{\min}$ adapts to these shifts while retaining the top SR. Under dataset transfer, our method remains strongest with SR \textbf{0.597} and SPL \textbf{0.465} (\textbf{+7.4}/\textbf{+11.6} over NavDP), preserving safer and more efficient navigation under domain shift.

\subsection{Sensitivity to ESDF Construction and Inference Cost}
\label{sec:sens_speed}
At inference, our generator--selector policy runs purely on RGB-D observations and goal information, without ESDF reconstruction or map building; ESDFs are used only offline to supervise the teacher. We report (i) sensitivity to ESDF construction hyperparameters and (ii) per-step inference cost of the generator and selector versus candidate set size $K$, on HM3D following Sec.~\ref{sec:simtest}. Our default ESDF uses a $5.0$\,cm voxel size and $0.4$\,m truncation band; for each setting we retrain teacher and student. We time forward passes on a laptop GPU (NVIDIA RTX 4060), excluding environment stepping and rendering, averaging over 200 iterations after 50 warm-ups with \texttt{torch.cuda.synchronize()} around each timed region. Results are in Table~\ref{tab:sens_speed}.

\begin{table}[t]
\centering
\caption{ESDF sensitivity (Panel A) and per-step inference efficiency with different $K$ (Panel B) on HM3D.}
\label{tab:sens_speed}

\setlength{\tabcolsep}{5.0pt}
\renewcommand{\arraystretch}{1.05}

\newlength{\panelwidth}
\setlength{\panelwidth}{\linewidth}

\begin{tabular*}{\panelwidth}{@{\extracolsep{\fill}}cccc@{}}
\toprule
\multicolumn{4}{c}{Panel A: ESDF Sensitivity (Default in First Line).} \\
\midrule
\multicolumn{1}{c}{Voxel (cm)} &
\multicolumn{1}{c}{Trunc (m)} &
\multicolumn{1}{c}{$\mathrm{SR}(\uparrow)$} &
\multicolumn{1}{c}{$\mathrm{SPL}(\uparrow)$} \\
\midrule
\multicolumn{1}{c}{5.0} & \multicolumn{1}{c}{0.4} & \multicolumn{1}{c}{0.783} & \multicolumn{1}{c}{0.611} \\
\multicolumn{1}{c}{2.5} & \multicolumn{1}{c}{0.4} & \multicolumn{1}{c}{0.807} & \multicolumn{1}{c}{0.612} \\
\multicolumn{1}{c}{7.5} & \multicolumn{1}{c}{0.4} & \multicolumn{1}{c}{0.747} & \multicolumn{1}{c}{0.607} \\
\multicolumn{1}{c}{5.0} & \multicolumn{1}{c}{0.2} & \multicolumn{1}{c}{0.740} & \multicolumn{1}{c}{0.598} \\
\multicolumn{1}{c}{5.0} & \multicolumn{1}{c}{0.6} & \multicolumn{1}{c}{0.753} & \multicolumn{1}{c}{0.626} \\
\bottomrule
\end{tabular*}

\begin{tabular*}{\panelwidth}{@{\extracolsep{\fill}}ccccc}
\toprule
\multicolumn{5}{c}{Panel B: Per-step Inference Cost (Default $K{=}16$).} \\
\midrule
\multicolumn{1}{c}{$K$} &
\multicolumn{1}{c}{$\mathrm{SR}(\uparrow)$} &
\multicolumn{1}{c}{$\mathrm{SPL}(\uparrow)$} &
\multicolumn{1}{c}{Gen(ms/step)$(\downarrow)$} &
\multicolumn{1}{c}{Sel(ms/step)$(\downarrow)$} \\
\midrule
\multicolumn{1}{c}{8}  & \multicolumn{1}{c}{0.717} & \multicolumn{1}{c}{0.601} & \multicolumn{1}{c}{38\,($\pm 2$)} & \multicolumn{1}{c}{16\,($\pm 2$)} \\
\multicolumn{1}{c}{16} & \multicolumn{1}{c}{0.783} & \multicolumn{1}{c}{0.611} & \multicolumn{1}{c}{56\,($\pm 2$)} & \multicolumn{1}{c}{21\,($\pm 2$)} \\
\multicolumn{1}{c}{32} & \multicolumn{1}{c}{0.790} & \multicolumn{1}{c}{0.620} & \multicolumn{1}{c}{94\,($\pm 2$)} & \multicolumn{1}{c}{28\,($\pm 2$)} \\
\bottomrule
\end{tabular*}
\end{table}

\subsection{Real-World Experiments}

We deploy our policy on a Unitree G1 humanoid with an Intel RealSense D435i RGB-D camera, without task-specific tuning, across three settings of increasing difficulty: an easy corridor with wide aisles, a medium apartment with tight doorways, and a hard cluttered laboratory with narrow passages and irregular obstacles. In each we run 10 episodes with targets sampled in reachable free space, counting success within the standard radius under a fixed timeout. We compare the same baselines, replacing iPlanner with EGO-Planner~\cite{zhou2020ego}, deployable without training. Results are in Table~\ref{tab:real}.

\begin{table}[t]
\centering
\caption{Real-world PointGoal results with 10 episodes per scene.}
\label{tab:real}
\setlength{\tabcolsep}{5.0pt}
\renewcommand{\arraystretch}{1.02}
\begin{tabular}{lccc}
\toprule
Method & Corridor (easy) & Apartment (medium) & Lab (hard) \\
\midrule
NavDP         &  10/10 (1.0) &  9/10 (0.9) &  6/10 (0.6) \\
ViPlanner      &  9/10 (0.9) & 6/10 (0.6) &  4/10 (0.4) \\
PointNav      &  5/10 (0.5) &  5/10 (0.5) &  0/10 (0.0) \\
EGO-Planner   &  5/10 (0.5) &  4/10 (0.4) &  1/10 (0.1) \\
\textbf{Ours} &  \textbf{10/10 (1.0)} &  \textbf{10/10 (1.0)} &  \textbf{8/10 (0.8)} \\
\bottomrule
\end{tabular}
\end{table}

\begin{figure}[t]
\centering
\includegraphics[width=\linewidth]{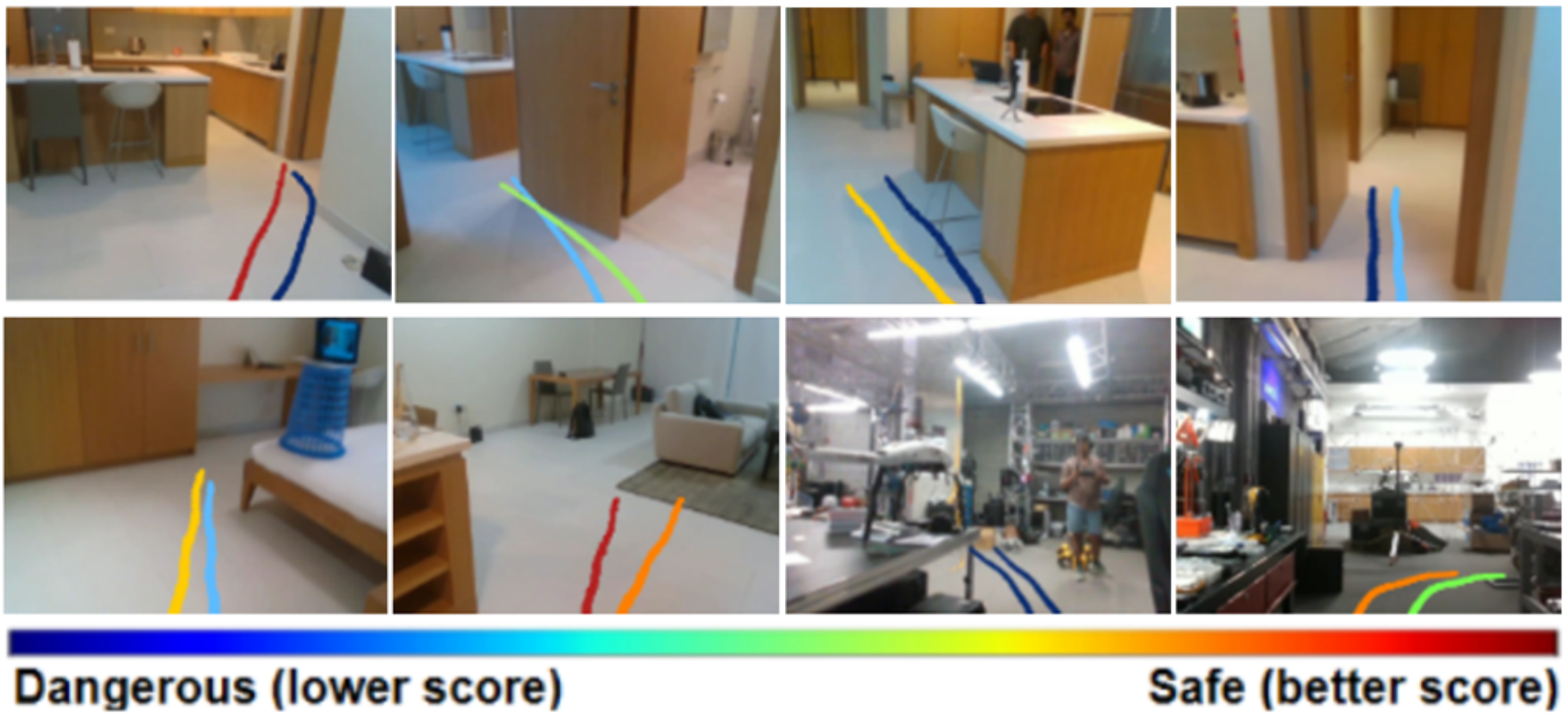}
\caption{Real world test result visualization. We project two trajectories with the highest score and the lowest score to the robot first-person perspective. The color from blue to red represents the score from low to high, which also means from dangerous to safe.}
\label{fig:grid9}
\end{figure}

Across these scenes the policy reliably reaches targets while respecting local kinematics. For qualitative inspection, we back-project the highest- and lowest-scored trajectories onto the current image with a blue-to-red colorbar for low-to-high critic score (Fig.~\ref{fig:grid9}). In all three environments, including the cluttered laboratory, our policy maintains stable progress and consistent safety margins, showing the learned critic generalizes across scene difficulty.

\begin{figure}[t]
  \centering
  \includegraphics[width=\linewidth]{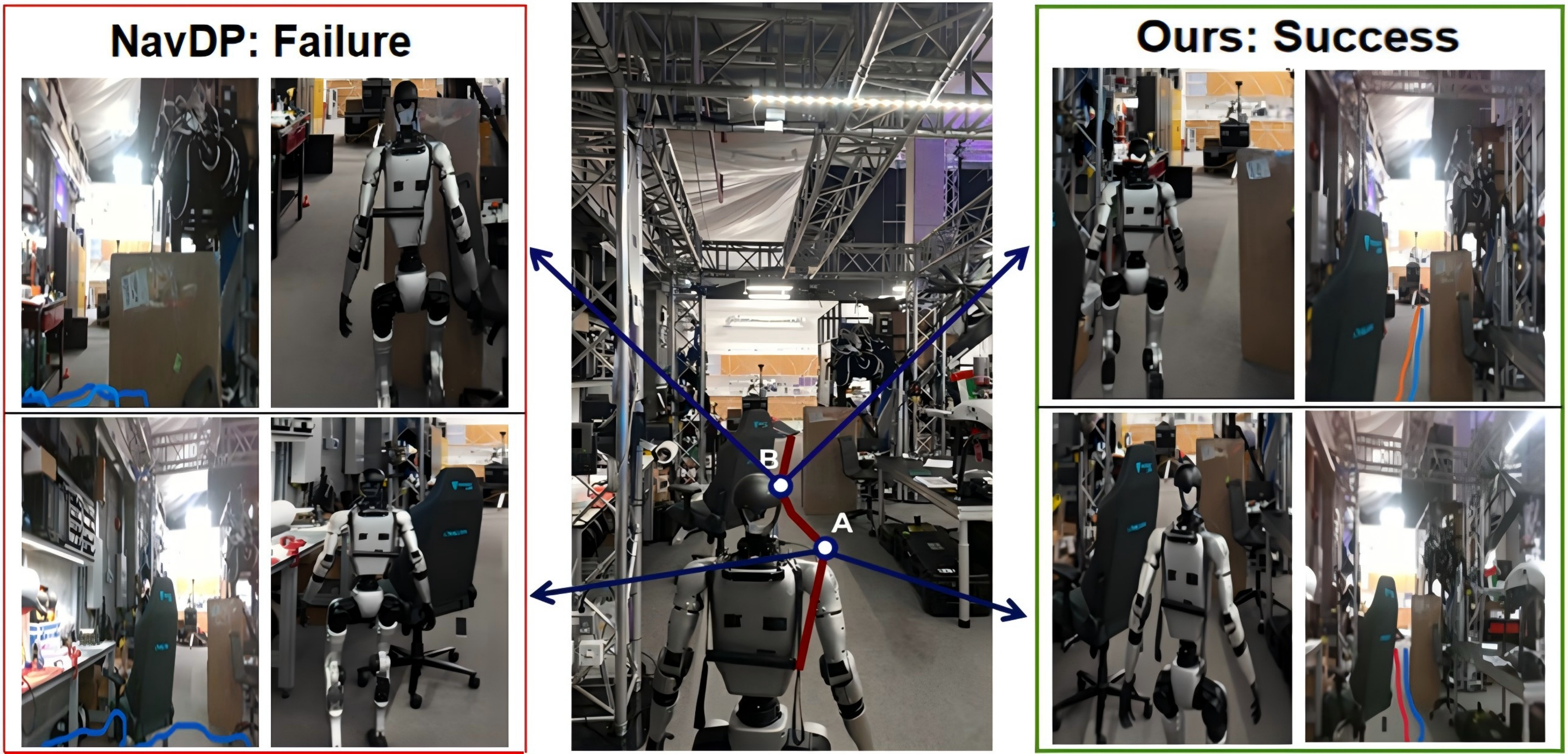}
  \caption{A real world test with check points A and B, both of which have the DoN close to 0.75. NavDP's policy failed in both check points while our policy successfully passes the cluttered obstacles. The left four photos are NavDP's failed cases with the third-person perspectives and visualization of trajectories, and the right four are our success cases with the third-person perspectives and visualization of trajectories.}
  \label{fig:nine-collage}
\end{figure}

To probe adaptability of the learned constraint, we use the Degree of Narrowness $\mathrm{DoN}=w_{\mathrm{robot}}/w_{\min}$, where \(w_{\mathrm{robot}}\) is the minimum safe width a robot can pass through and \(w_{\min}\) is the minimum passable space width; a larger DoN (closer to \(1\)) indicates a narrower, harder condition.
We construct two test points $A$ and $B$ where detours are impossible, so the only feasible outputs are short left/right shaking or stopping (Fig.~\ref{fig:nine-collage}). At \(\mathrm{DoN}{=}0.60\) both methods pass. At \(\mathrm{DoN}{=}0.75\) our policy still passes by tightening its learned constraint while keeping clearance above the budget, whereas NavDP with a fixed \(d_{\min}{=}0.5\,\mathrm{m}\) judges the corridor non-passable and stalls. At \(\mathrm{DoN}{=}0.90\) neither passes: clearance falls below a safe envelope and our critic refuses to further relax \(d_{\min}\), showing the learned margin balances adaptability with safety rather than shrinking below a physically meaningful floor.

\subsection{Ablation: Safety-Critic Structure}
\label{sec:ablation}
We conduct four ablation tests on HM3D with 300 episodes and the same start--goal pairs across all runs, so behavior changes reflect the safety-critic structure rather than noise. All models share the same generator; only the critic differs. We remove one component from~\eqref{eq:v_ours_sum}: \textbf{Model A}---remove the safety term~\eqref{eq:v_safe}; \textbf{Model B}---remove the efficiency term~\eqref{eq:v_efficient}; \textbf{Model C}---remove the balancing term~\eqref{eq:v_balance}.

Since all tests share the same 300 pairs, beyond Successes, SR, and SPL we report a success-set overlap metric quantifying how much one variant's success set covers another (Table~\ref{tab:ablate_overlap}):
\[
\text{Overlap}(X\!\to\!Y)=\frac{\{\text{episodes solved by both }X\text{ and }Y\}}{\{\text{episodes solved by }Y\}}.
\]

\begin{table}[t]
\centering
\caption{Ablation tests on HM3D with 300 episodes.}
\label{tab:ablate_main}
\setlength{\tabcolsep}{10.0pt}
\renewcommand{\arraystretch}{1.05}
\begin{tabular}{lccc}
\toprule
Model & Successes & SR $\uparrow$ & SPL $\uparrow$ \\
\midrule
removing $V_{\mathrm{safe}}$ (A)       & 207/300 & 0.690 & 0.572 \\
removing $V_{\mathrm{efficient}}$ (B)  & 212/300 & 0.707 & 0.468 \\
removing $V_{\mathrm{balance}}$ (C)    & 190/300 & 0.633 & 0.585 \\
\textbf{Ours}                          & \textbf{235/300} & \textbf{0.783} & \textbf{0.611} \\
\bottomrule
\end{tabular}
\end{table}


\begin{table}[t]
\centering
\caption{Success-set overlap.
$\text{Overlap}(X\!\to\!Y)=\frac{\{\text{episodes solved by both }X\text{ and }Y\}}{\{\text{episodes solved by }Y\}}.$}
\label{tab:ablate_overlap}
\setlength{\tabcolsep}{7.5pt}
\renewcommand{\arraystretch}{1.05}
\begin{tabular}{lcc}
\toprule
Model & Overlap w/ A $\uparrow$ & Overlap w/ Ours $\uparrow$ \\
\midrule
removing $V_{\mathrm{safe}}$ (A)      & ---                       & $197/235=0.838$ \\
removing $V_{\mathrm{efficient}}$ (B) & $167/207=0.821$           & $180/235=0.766$ \\
removing $V_{\mathrm{balance}}$ (C)   & $147/207=0.720$           & $161/235=0.685$ \\
\textbf{Ours}                         & $197/207=0.952$           & ---             \\
\bottomrule
\end{tabular}
\end{table}

Removing $V_{\mathrm{efficient}}$ raises SR over removing $V_{\mathrm{safe}}$ (0.690 to 0.707) but drops SPL to 0.468, as without an explicit efficiency objective the selector over-selects conservative detours and times out. Removing $V_{\mathrm{balance}}$ shows the opposite trend (SR 0.633, SPL 0.585): more direct but more failure-prone selections in narrow areas, since the budget--clearance anchor is lost and miscalibration distorts the gate and remaining penalties. Removing $V_{\mathrm{safe}}$ (SR 0.690, SPL 0.572) retains the gate and coupling but lacks explicit pointwise and transition-wise safety regularization, making it vulnerable in tight near-contact regimes.

The full model achieves the best SR 0.783 and SPL 0.611: $V_{\mathrm{safe}}$ filters risky proposals, $V_{\mathrm{efficient}}$ reduces detour-driven timeouts under a safety-aware gate, and $V_{\mathrm{balance}}$ stabilizes the budget so higher scores reflect real clearance gains. The overlap confirms ours covers 197 of A's 207 successes and adds 38 new ones, while C's lower overlap marks a distinct, efficiency-seeking subset that is less reliable under clutter. All three terms are thus required for a well-calibrated selector.

\section{Conclusion}
We introduced a generate--select framework that replaces fixed margins with a learnable, context-conditioned safety critic fusing safety and efficiency in balance. Across simulation and real-world tests it achieves the highest SR and SPL, and transfers from pure simulation to a real humanoid in cluttered indoor scenes without task-specific tuning. Ablations confirm each critic term contributes to a favorable safety--efficiency trade-off and consistent gains over fixed-threshold baselines.

\section*{ACKNOWLEDGMENT}
Authors appreciate the support provided by the NYUAD Center for Artificial Intelligence and Robotics (CAIR), funded by Tamkeen under the NYUAD Research Institute Award CG010.



{
    \small
    \bibliographystyle{IEEEtran}
    \bibliography{refs}
}
\end{document}